\documentclass[nohyperref]{article}

\usepackage{microtype}
\usepackage{graphicx}
\usepackage{subfigure}
\usepackage{booktabs} 
\usepackage{array}

\newcolumntype{P}[1]{>{\raggedright\arraybackslash}p{#1}}
\usepackage{multirow}

\usepackage{hyperref}

\usepackage{natbib}
\setcitestyle{numbers}
\setcitestyle{square}

\usepackage[accepted]{icml2022}

\usepackage{amsmath}
\usepackage{amssymb}
\usepackage{mathtools}
\usepackage{amsthm}

\usepackage[capitalize,noabbrev]{cleveref}

\theoremstyle{plain}

\theoremstyle{definition}

\theoremstyle{remark}

\icmltitlerunning{Improved Text Classification Accuracy via TTA}

\begin{document}

\twocolumn[
\icmltitle{Improved Text Classification via Test-Time Augmentation}



\icmlsetsymbol{equal}{*}

\begin{icmlauthorlist}
\icmlauthor{Helen Lu}{yyy}
\icmlauthor{Divya Shanmugam}{yyy}
\icmlauthor{Harini Suresh}{yyy}
\icmlauthor{John Guttag}{yyy}
\end{icmlauthorlist}

\icmlaffiliation{yyy}{Computer Science and Artificial Intelligence Laboratory (CSAIL), Massachusetts Institute of Technology (MIT), Cambridge, MA, USA}

\icmlcorrespondingauthor{Helen Lu}{helenl@mit.edu}
\icmlcorrespondingauthor{Divya Shanmugam}{divyas@mit.edu}
\icmlcorrespondingauthor{Harini Suresh}{hsuresh@mit.edu}
\icmlcorrespondingauthor{John Guttag}{guttag@csail.mit.edu}

\icmlkeywords{Machine Learning, Natural Language Processing, Data Augmentation}

\vskip 0.3in
]



\printAffiliationsAndNotice{} 

\begin{abstract}
Test-time augmentation---the aggregation of predictions across transformed examples of test inputs---is an established technique to improve the performance of image classification models. Importantly, TTA can be used to improve model performance post-hoc, without additional training. Although test-time augmentation (TTA) can be applied to any data modality, it has seen limited adoption in NLP due in part to the difficulty of identifying label-preserving transformations. In this paper, we present augmentation policies that yield significant accuracy improvements with language models. A key finding is that augmentation policy design–for instance, the number of samples generated from a single, non-deterministic augmentation–has a considerable impact on the benefit  of TTA. Experiments across a binary classification task and dataset show that test-time augmentation can deliver consistent improvements over current state-of-the-art approaches. 
\end{abstract}

\section{Introduction}
\label{introduction}

Designing accurate and robust machine learning models for classification tasks typically requires large labeled datasets. One standard practice for artificially expanding a given dataset is data augmentation, in which new data are generated by transforming existing examples. While data augmentation is often  performed during model training, recent work has shown that the use of data augmentation at inference, or test-time augmentation (TTA), can improve model accuracy \cite{krizhevsky2012imagenet, szegedy2015going, matsunaga2017image}, robustness \cite{prakash2018deflecting, song2017pixeldefend, cohen2019certified}, and uncertainty estimates \cite{ayhan2018test, matsunaga2017image, smith2018understanding, wang2019aleatoric}. TTA produces a final prediction by aggregating model predictions (usually by averaging) made from multiple transformed versions of a given test input. In addition to improved model performance, TTA is popular because of its low implementation burden—predictions use a pre-trained model, so no hyperparameter tuning is necessary, and inference on multiple examples can be easily parallelized. Leveraged successfully, TTA can improve model performance without requiring additional data or changes to the underlying model.

\begin{figure*}[ht]
\vskip 0.2in
\begin{center}
\centerline{\includegraphics[width=.8\textwidth]{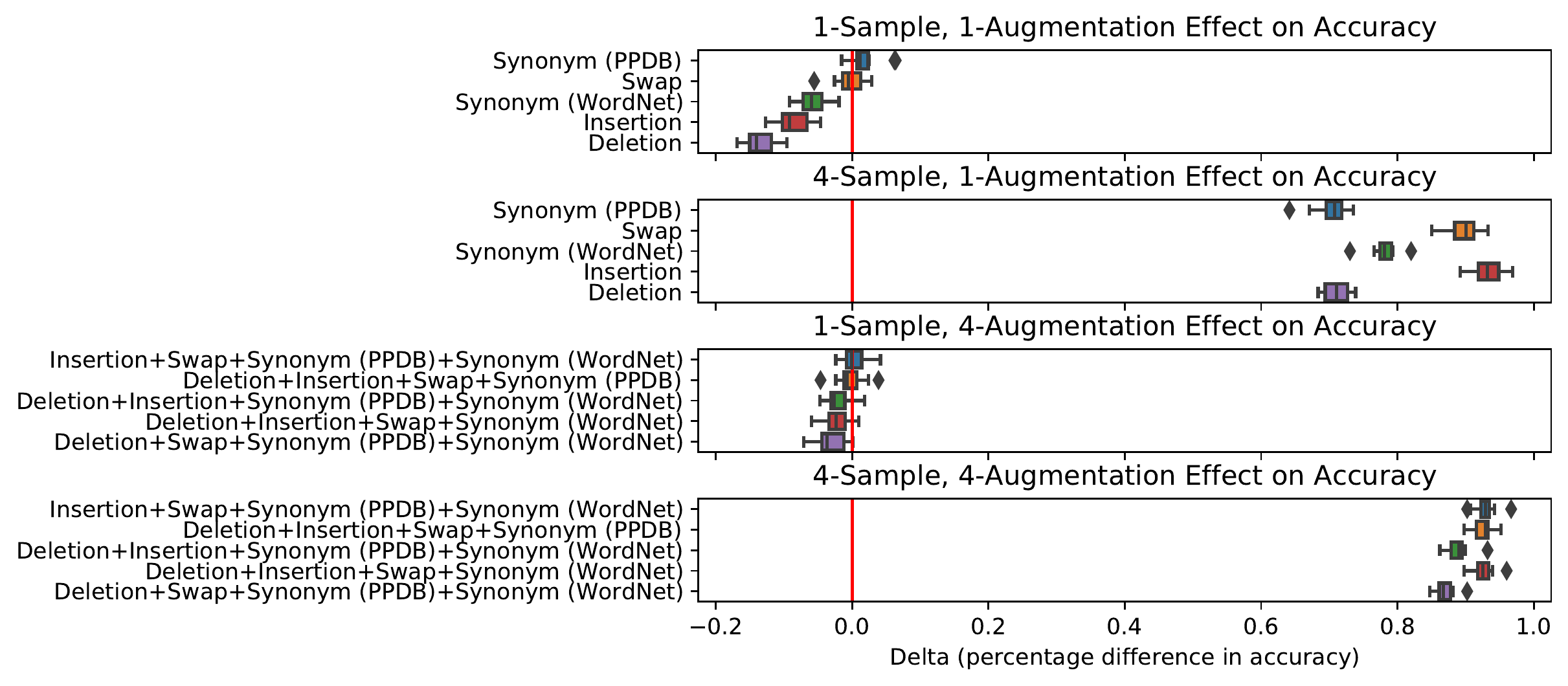}}
\caption{Impact of different augmentation policies on binary classification accuracy. The overall best-performing transformations are depicted and compared in each policy configuration. The change in accuracy is displayed in percentage points compared to the baseline, in which no augmentations were applied to inputs fed to the trained model. Accuracy was calculated by averaging across 10 random subsamples, each containing a random 80\% of the test inputs.}
\label{cc_stack}
\end{center}
\vskip -0.2in
\end{figure*}

While TTA is commonly used in computer vision (CV), there is relatively little research on the potential of TTA in natural language processing (NLP). One reason for this is that it is not clear which augmentations to apply to text inputs at test-time. CV enjoys a set of “standard” test-time augmentations, including flips, crops, and scales, which are often assumed to be label preserving. In contrast, label-preserving transformations to text are frequently task-dependent, and as a result, the design of performant test-time augmentation policies is not straightforward. A number of design choices, including which augmentations to choose and how to aggregate the resulting predictions, are involved in TTA. We focus on the first step–understanding which augmentations to choose–in the context of natural language processing. 

The goal of our work is to explore the use of TTA for improving NLP-based classification First we establish four types of TTA policies, apply these to text, and use a simple average to aggregate the resulting predictions. We then use this method to improve accuracy on binary classification of the WILDS CivilComments dataset. We empirically analyze the four different configurations and propose optimal augmentation policies that most significantly improve classification accuracy.

Our results establish that TTA can provide significant improvement to text classification accuracy.  We find that the scale of TTA accuracy increase is heavily dependent on augmentation policy design. \cref{cc_stack} compares the effect of four different augmentation policy configurations on classification accuracy. We determine that configurations aggregating across multiple samples from non-deterministic augmentations yields the greatest benefit and build intuition as to why this design significantly outperforms configurations using a single sample. We conclude with practical recommendations for the application of TTA in NLP. 

\section{Related Work}
\label{related_work}

Text augmentation at test-time has been applied to several tasks, including abstractive summarization \cite{fadaee-etal-2017-data, parida-motlicek-2019-abstract, zhu-2019}, question answering \cite{longpre-etal-2019-exploration, yang-2019, singh2019xlda}, sequence tagging \cite{ding2020daga, csahin2019data, dai2020analysis}, parsing \cite{jia2016data, yu2020grappa, andreas2019good}, grammatical error correction \cite{boyd2018using, zhang2019sequence, yang2022controllable}, and neural machine translation \cite{sennrich2015improving, fadaee-etal-2017-data, xia2019generalized}. Because our focus is test-time augmentation for the purpose of improving language model classification accuracy, we limit our discussion to work considering this problem. 

Our work focuses on augmentation policy design choices and why specific policies outperform others.

\section{Method}
\label{method}

We borrow terminology from \citet{shanmugam2021better} and assume two inputs to our method:

\begin{enumerate}
    \item A pre-trained black-box classifier $f : X \rightarrow R^C$ that maps text inputs to a vector of class logit predictions. We use $X$ to denote the space of inputs on which the classifier can operate and $C$ to denote the number of classes. We assume that $f$ is not fully invariant with respect to augmentations (i.e., predictions on transformations of a test input are not guaranteed to be the same).
    
    \item A list of $M$ augmentation transforms, $[a_1, … a_m]$ which we deem an augmentation policy $A$. Each transform $a_m : X \rightarrow X$ is a stochastic transform that modifies the given input at a character- or word-level based on parameters defining the probability the transform is applied and the magnitude of transformation (e.g. percent of words changed, percent of characters changed), where larger magnitudes translate to larger changes to the example.  
    
\end{enumerate}

Given a text input $t$, we apply an augmentation policy $A$ containing $M$ augmentation transforms to generate $M$ transformed inputs. All $M + 1$ inputs (transformed and the original) are then passed into the pre-trained classifier $f$ to generate $(M + 1)$ $R^C$ vectors containing the class logit predictions. We then generate a single prediction by applying a simple average to the $M + 1$ logit predictions. While there are more complex means of aggregating predictions that may maximize prediction performance, we choose to use averaging because it is the simplest version of TTA and suits our goal of understanding the baseline value of TTA in NLP.

\section{Experimental Set-Up}
\label{experiment}

We evaluate the performance of our method across a dataset and model architecture laid out in the following sections. 

\paragraph{Datasets}

The WILDS CivilComments dataset \cite{koh2021wilds}, a modification of the original CivilComments dataset \cite{borkan2019nuanced}, consists of 448,000 comments made on online articles. Each comment is labeled as either toxic or non-toxic using a binary indicator. Further details on the dataset can be found in \cref{dataset_summary}. We use the pre-defined test set to report accuracy because the default TTA approach to aggregation (i.e. averaging) requires no training data, and assumes access to a pre-trained model.

\paragraph{Augmentation Policies}
We consider two classes of augmentation policies from the \textit{nlpaug} library, consisting of fifteen different augmentation transformations. Insertion, deletion, and substitution of randomly-selected characters are categorized as \textit{character-based} and of randomly-selected words as \textit{word-based}. 

For each class of augmentation policies, we chose parameters that maximized accuracy on a held-out validation set. Character-based augmentation policies transformed a random character from 10\% of randomly-selected words of an input, with the constraint that any selected word less than 4 characters long would remain unmodified. Word-based augmentation policies modified one random word from a given input, and sentence-based augmentation policies use the default parameters of the \textit{nlpaug} library. 

\paragraph{Models}
For all experiments, we used a pre-trained network made available by Koh et al.\yrcite{koh2021wilds}. Hyperparameters and training details can be found in \cref{model_hyperparameters}. The predictions generated from these models on unaugmented test inputs served as our baseline accuracy. 

\paragraph{Augmentation Configurations}

Augmentation policies can vary in terms of the type of augmentations, and the number of samples drawn from a particular augmentation. Using these two dimensions, we constructed four representative test-time augmentation policy configurations and compared their effects on classification accuracy. The configurations are as follows:

\textit{1-sample, 1-augmentation}: Averages predictions over the original, unaugmented input and predictions generated following augmentation policy $A = [a_m]$.

\textit{1-sample, 4-augmentations}: Averages predictions over the original, unaugmented input and four prediction vectors, each generated by a different augmentation transform in policy $A = [a_m , a_n , a_p, a_q]$. All transforms were of the same class (i.e. either all word-based or all character-based).

\textit{4-samples, 1-augmentation}: Averages predictions over the original, unaugmented input and four prediction vectors, each generated by the same augmentation transform $A = [a_m, a_m, a_m, a_m]$. 

\textit{4-samples, 4-augmentations}: Averages of the class logit prediction vector of the original, unaugmented input and sixteen class logit prediction vectors: four samples generated by each of four different augmentation transforms $a_m$ , $a_n$ , $a_p$, and $a_q$. All transforms were of the same class (i.e. either all word-based or all character-based).

Since each augmentation transform (e.g., word deletion) is stochastic, multiple augmented samples were created. Specifically four samples were generated to maximize accuracy based on analysis with a held-out validation set. Policies aggregating four distinct augmentation transforms were designed to complement the optimal four sample configurations outlined above.

\paragraph{Statistical Significance}
We use a pairwise t-test to measure the statistical significance (with respect to the baseline model) of our accuracy results, which were obtained by averaging 10 random subsamples containing 80\% of the test set. 

\section{Results}
\label{results}

Our results demonstrate that the multi-sample single augmentation configuration performs best. We walk through each configuration's results in detail below. We focus on word-based augmentation transforms in the following sections after finding that character-based augmentation transforms were consistently unsuccessful in yielding accuracy improvement.

For the 1-sample, 1-augmentation configuration, only the augmentation policy using the transform \textit{ppdb\_synonym} (substitution of randomly chosen words with a synonym determined by the PPDB database) outperformed the baseline model’s classifications on average, as shown in \cref{cc_stack}. However, the increase in accuracy was found to be statistically insignificant. 

For the 4-samples, 1-augmentation configuration, all nine word-based policies yielded statistically-significant improvements to classification accuracy, from 0.6\% to 0.9\%.  (\cref{cc_stack}). 

For the 1-sample, 4-augmentations configuration, four augmentation policies slightly outperformed the baseline model’s classifications, but not significantly (\cref{cc_stack}). All four augmentation policies consisted of four distinct word-based transforms. 

For the 4-samples, 4-augmentations configuration, all 126 augmentation policies–combinations of four out of nine word-based augmentation transforms–significantly outperformed the baseline model’s classifications, yielding improvements from 0.8\% to 1.0\% (\cref{cc_stack}). 

\subsection{Comparing 4-Sample and 1-Sample Policies}

\begin{figure}[t!]
\vskip 0.2in
\begin{center}
\centerline{\includegraphics[width=\columnwidth]{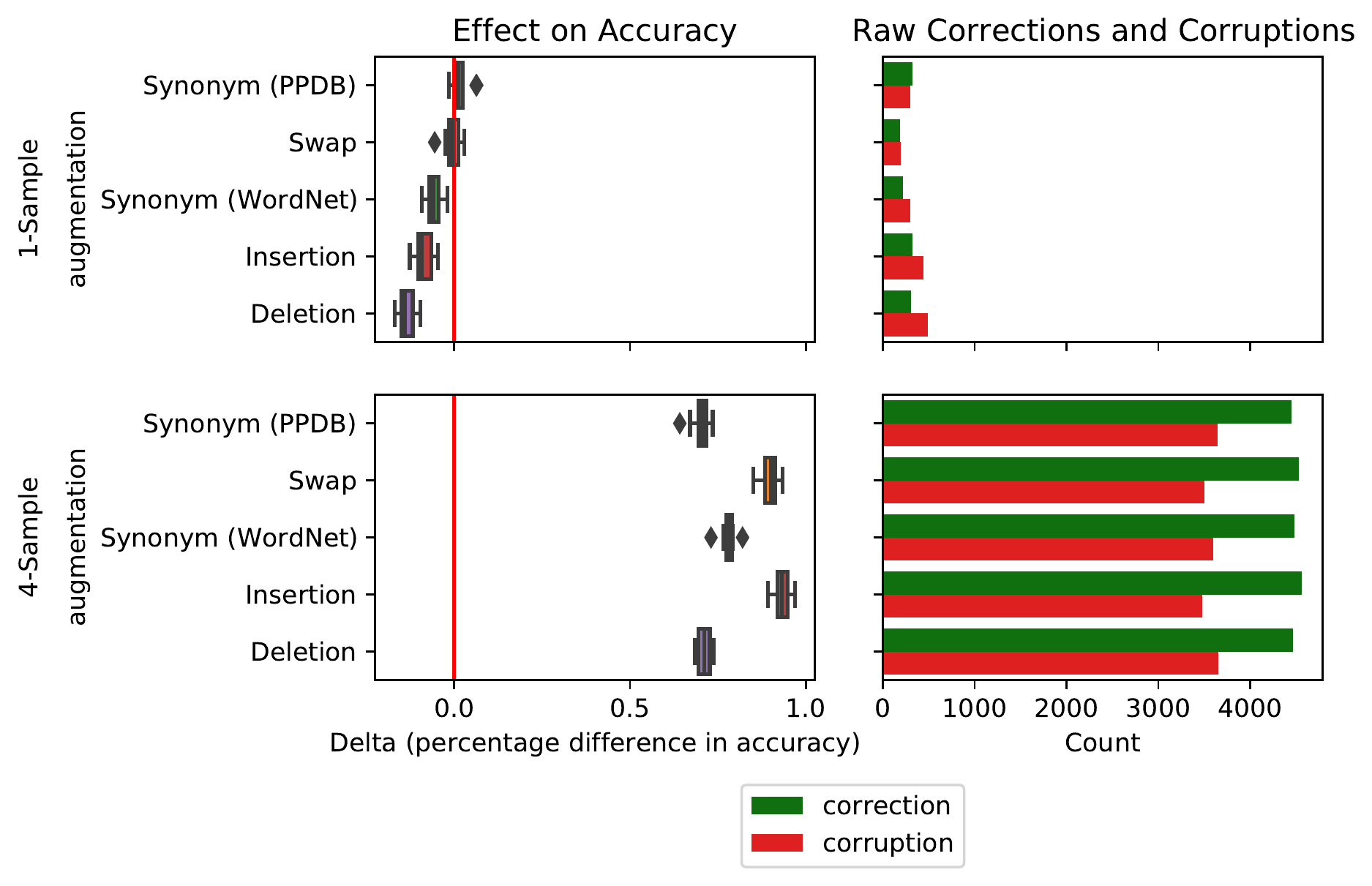}}
\caption{Corrections and corruptions of 1-sample, 1-augmentation and 4-samples, 1-augmentation policies. The change in accuracy is in percentage points compared to the baseline. Accuracy was calculated by averaging across 10 samples, each containing a random 80\% of the test inputs.}
\label{cc_corrections_corruptions_count}
\end{center}
\vskip -0.2in
\end{figure}

Our results show that multi-sample single-augmentation TTA configurations offer significant improvement in accuracy, while single-sample single-augmentation configurations do not. This is surprising, because any individual augmentation \emph{decreases} the pre-trained model's classification accuracy on average (Fig. \ref{cc_corrections_corruptions_count}), where each augmentation introduces more corruptions (red) than corrections (green). Here, we explore the mechanism that drives this difference in performance.

We first examined the count of corrections (i.e. inputs initially misclassified by the pre-trained model but correctly classified after applying an augmentation transform) and corruptions (i.e. inputs initially correctly classified by the pre-trained model but misclassified after applying an augmentation transform). \cref{cc_corrections_corruptions_count} shows that that the 4-sample, 1-augmentation policy changes many more examples than the 1-sample policy. Importantly, it consistently creates more corrections than corruptions, leading to an overall gain in accuracy. 

Why do multiple samples lead to more corrections than corruptions? We hypothesized that for corrections, there may be a greater overlap between individual samples\,---\,i.e., when an augmentation improves the model's prediction, multiple samples are likely to agree, and their average will result in a strong correction. On the other hand, corruptions may be more randomly distributed\,---\,and averaging across them washes out any individual decrease in accuracy. We tested this hypothesis by calculating how often sample predictions overlapped for corrections vs corruptions. In line with our hypothesis, we found that for corrected examples, there was an average overlap of around 0.51 across samples, as compared to only 0.23 for corrupted examples (see \cref{overlap_supporting_figure} for supporting figure).

\section{Discussion}
\label{discussion}

In this paper, we investigate accuracy improvements when TTA is combined with a large pre-trained language model. Through an analysis of the WILDS CivilComments dataset, we show that TTA can introduce significant improvements in classification accuracy.  

Our results show that policy design significantly influences accuracy improvements of TTA. Fixing the number of augmentations in the policy, we compare results using two approaches: one which produces multiple samples from a single augmentation, and another which uses single samples from different augmentations. We find that drawing multiple samples from a single transform produces significantly larger improvements in accuracy: from 0.06\% to 1\%. 

A natural extension to the multi-sample single-augmentation approach is to replicate it with multiple augmentations. This produces the multi-sample multi-augmentation configuration and we find that it yields a marginal, often insignificant benefit (at most a .04\% improvement in accuracy). Given how the multi-augmentation multi-sample configuration comes with the computational cost of many more predictions at test-time, this suggests that multi-sample single-augmentation approaches offer a better tradeoff between computational efficiency and accuracy.

We also observe that word-based transforms consistently outperform character-based transforms. This is likely due to how word-based transforms modify inputs. The best-performing word-based transforms incorporated semantic meaning via predefined libraries (WordNet Synonym, PPDB Synonym) or embedding models. It is likely that the alterations made by character-based transforms, on the other hand, are altering fundamental word stems. These stems are often used by natural language models to encode words (and their meanings). Modifying a single character may generate a new stem with unrelated (or opposite) meaning, resulting in misclassification of the transformed input. 

The accuracy improvements generated from applying TTA are significant for three key reasons. First, TTA can outperform a model fine-tuned to perform well on test data. Despite the fact that the original accuracy is 92.3\%, TTA can offer improvements of up to 93.7\%. Secondly, the baseline model was trained without training-time augmentation \cite{koh2021wilds}. This indicates that TTA may have even greater potential in NLP contexts than in CV, where augmentations used in test-time often parallel those used by the model in train-time. Thirdly, our approach yielded improvements even when using a simple aggregation scheme of averaging. It is likely that, similar to computer vision, greater improvements are possible by weighting augmentations differently.

\nocite{fang2021survey}

\bibliography{paper}
\bibliographystyle{icml2022}

\newpage
\appendix
\onecolumn
\section{CivilComments Dataset}
\label{dataset_summary}

The WILDS CivilComments dataset (Koh et. al 2021), a modification of the original CivilComments dataset (Borkan et. al., 2019), consists of 448,000 comments made on online articles. Ranging from word lengths of 1 to 315, the comments are labeled as either toxic or non-toxic using a binary indicator obtained from crowdsourcing and majority voting among at least 10 crowdworkers. Each comment is also annotated with binary indicators regarding content mention of nine demographic identities (male, female, LGBTQ, Christian, Muslim, other religions, black, white, other) and six characteristics (severely toxic, obscene, threatening, insulting, identity attacking, and sexually explicit). See \cref{civilcomments_dataset_sample} for examples of a toxic and non-toxic comment.

\begin{table}[t]
\caption{Sample of Toxic and Non-Toxic Comment from the WILDS CivilComments dataset.}
\label{civilcomments_dataset_sample}
\vskip 0.15in
\begin{center}
\begin{small}
\begin{sc}
\begin{tabular}{>{\centering\arraybackslash}p{2cm} P{7cm} >{\centering\arraybackslash}p{1cm} >{\centering\arraybackslash}p{1cm} >{\centering\arraybackslash}p{1cm} >{\centering\arraybackslash}p{2cm}}
\toprule
Toxicity & Comment & Male & Female & ... & Identity Attack \\
\midrule
0 & No, he was accused of being a racist white man. & 1 & 0 & ... & 0 \\
1 & The truth hurts so you have to reply with your idiotic comments. You really do not realize how ignorant you are. & 0 & 0 & ... & 0 \\

\bottomrule
\end{tabular}
\end{sc}
\end{small}
\end{center}
\vskip -0.1in
\end{table}

\section{Model Hyperparameter Settings}
\label{model_hyperparameters}
For all experiments, we used a state-of-the-art classifier developed for the WILDS CivilComments. We downloaded the pre-trained models, which were fine-tuned DistilBERT-base-uncased models using implementation from Sanh et al.\cite{wolf}  and hyperparameter settings from Koh et al \cite{koh2021wilds}: batch size $16$; learning rate $10^{-5}$ using the AdamW optimizer \cite{AdamW} for $5$ epochs with early stopping; an L2-regularization strength of $10^{-2}$; a maximum number of tokens of $300$; a learning rate chosen through a grid search over ${10^{-6}, 2 \times 10^{-6}, 10^{-6}, 2 \times 10^{-5}}$; and default values for all other hyperparameters. 

\section{4-Sample Corrections Corruptions Analysis}
\label{overlap_supporting_figure}

\begin{figure}[ht]
\begin{center}
\centerline{\includegraphics[scale = 0.7]{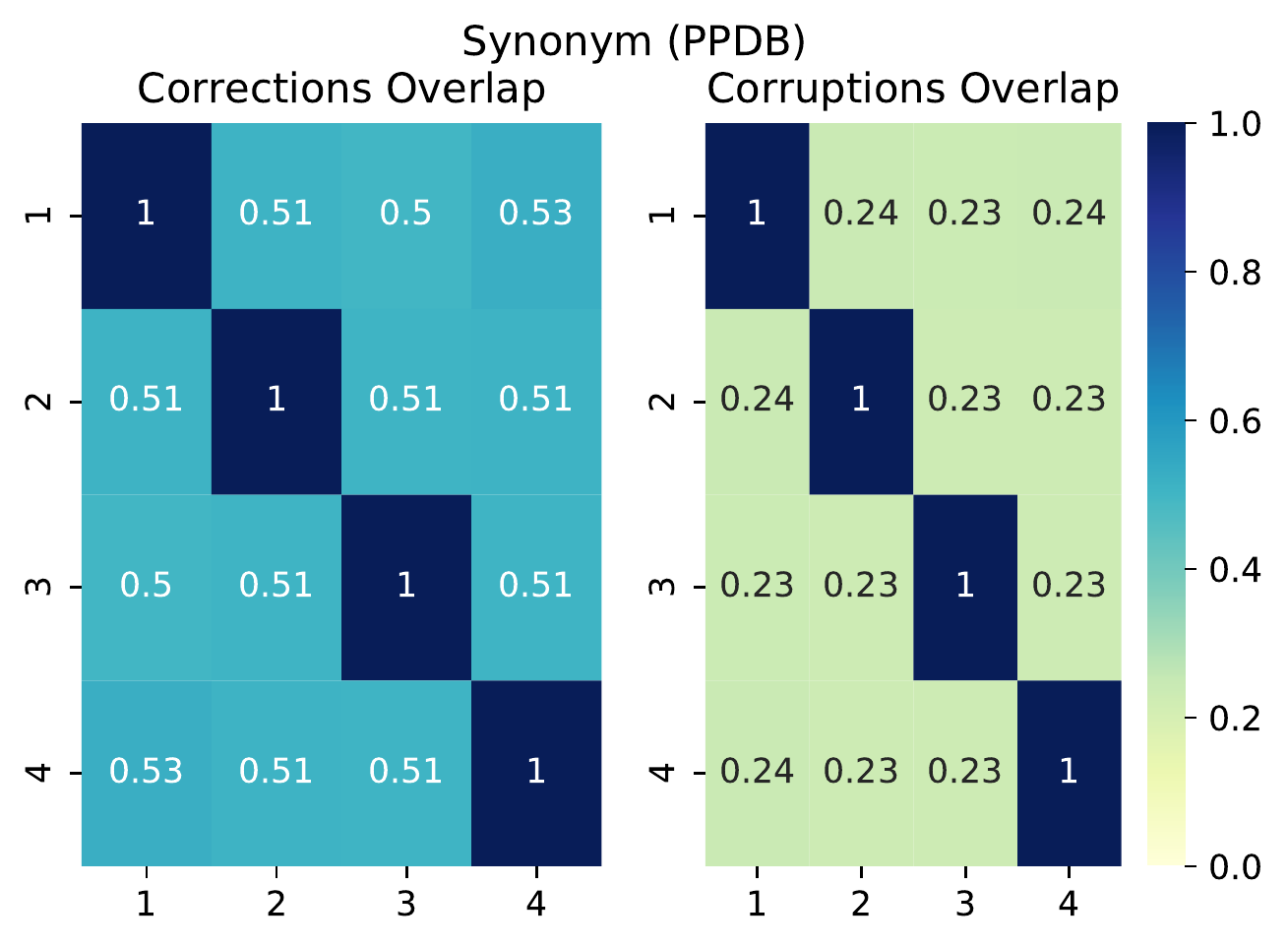}}
\caption{Fractional overlap of corrections and corruptions across pairs of samples generated by applying the 4-sample, 1-augmentation policy with the Synonym (PPDB) transform. For two samples (without loss of generality, Sample 1 and Sample 2), overlap was calculated by taking the results from dividing the number of inputs that were corrected/corrupted by both Sample 1 and Sample 2 by the total unique inputs that were corrected/corrupted by either Sample 1 or Sample 2. The Synonym (PPDB) transform was selected without loss of generality.}
\label{cc_sample_level_heatmap}
\end{center}
\vskip -0.2in
\end{figure}

\end{document}